\documentclass[twocolumn,10pt]{IEEEtran}

\usepackage{multirow}
\usepackage[pdftex]{graphicx}
\usepackage{amsfonts,amsmath,amssymb,amsthm,epsfig,subfigure}
\usepackage{epsfig}

\newcommand{\be}{\begin{eqnarray}}
\newcommand{\ee}{\end{eqnarray}}

\def\bfY{{\bf Y}}
\def\bfhY{{\bf\hat{Y}}}

\def\bftY{{\bf \tilde{Y}}}
\def\hy{\hat{y}}
\def\hY{\hat{Y}}
\def\ty{\tilde{y}}

\def\tmu{\tilde{\mu}}
\def\tvar{\tilde{\var}}
\def\var{\mbox{var}}

\def\eg{{\em e.g.},~}
\def\ie{{\em i.e.},~}
\def\cf{{\em cf.},~}
\def\E{{\sf E}}
\def\P{{\sf P}}
\def \ra {\rightarrow}

\newcommand{\ben}{\begin{enumerate}}
\newcommand{\een}{\end{enumerate}}
\newcommand{\beq}{\begin{equation}}
\newcommand{\eeq}{\end{equation}}
\newcommand{\beqa}{\begin{eqnarray*}}
\newcommand{\eeqa}{\end{eqnarray*}}
\newcommand{\bit}{\begin{itemize}}
\newcommand{\eit}{\end{itemize}}
\newcommand{\bt}{\begin{tabular}{c}}
\newcommand{\btt}{\begin{tabular}}
\newcommand{\et}{\end{tabular}}

\newtheorem{proposition}{Proposition}

\newtheorem{lemma}{Lemma}

\newtheorem{corollary}{Corollary}

\begin{document}
\bibliographystyle{plain}

\title{A Study of Unsupervised
Adaptive
Crowdsourcing\thanks{This work is supported in part by NSF CNS grant 0916179.}}

\author{\authorblockN{G. Kesidi$\mbox{s}^{1,2}$ and A. Kurv$\mbox{e}^2$} \\
\authorblockA{${}^{1}$CS\&E Dept and ${}^2$EE Dept \\
The Pennsylvania State University, University Park, PA, 16802}\\
\{gik2 and ack205\}@psu.edu
}

\maketitle

\begin{abstract}
We consider
unsupervised crowdsourcing performance
based on the model given in \cite{Karger11}
wherein the responses of end-users are essentially rated
according to how their responses correlate with the majority of other
responses to the same subtasks/questions.
In one setting, we consider an independent sequence of
identically distributed crowdsourcing assignments (meta-tasks),
while in the other we consider a single assignment
with a large number of component subtasks.
Both problems yield intuitive results in which the
overall reliability of the crowd is a factor.
\end{abstract}


\section{Introduction}

On-line crowdsourcing addresses the problem of  solving a
large meta-task by decomposing it into a large number of
small tasks/questions and assigning them to an online community
of peers/users. Examples of decomposable meta-tasks
include  \cite{BerkeleySeminar,Hoffman11}:
\begin{itemize}
\item annotating (including recommending) or classifying a large number of
consumer products and services, or data objects
such as documents \cite{IPW10}, web sites (\eg answering which among a large
body of URLs contains pornography), images, videos;
\item  translating or transcribing a document \cite{Cas} possibly
including decoding a body of CAPTCHAs \cite{Savage10};
\item document correction through proofreading \cite{BLM10,Soy}; and
\item creating and maintaining content, \eg Wikipedia and open-source
communities.
\end{itemize}
General purpose platforms for on-line crowdsourcing
include Amazon's Mechanical Turk \cite{Mec,IPW10,Kulk11}
and Crowd Flower \cite{Cro}.

Users responding to questions may do so with different degrees of reliability.
If $p$ is the probability that a user correctly answers a question,
let the expectation $\E p$ be taken
over the ensemble of users.
Thus, $\E p$ is a measure of
the reliability of the {\em majority}
and, fundamentally, whether the positive correlation with the majority
ought to be sought for individual users
(as is typically assumed in many online
unsupervised ``polling" systems).

A user population is arguably
reliable ($\E p > 0.5$) when the population {\em itself} ultimately
decides the issue (\eg confidence intervals for an election poll),
or the questions concern a commonplace issue with commonplace
expertise among the population (\eg whether a web site contains pornography),
or the population is significantly financially incentivized to be accurate
(\ie incentivized to acquire the required expertise to be accurate).
Some market-based crowdsourcing scenarios (\eg questions of investing
in stocks of complex companies), or analogies to
bookmaking (setting odds so that the house always profits), may
not be relevant here, \ie  scenarios
where questions are pushed to users  who
minimally profit by answering them correctly.
That is,
for some specialized technical issues, it may be possible that the ``crowd"
will be unhelpful ($\E p \approx 0.5$) or incorrectly prejudiced/biased
($\E p < 0.5$).
In many cases, the users may need to be
 paid for questions answered \cite{Singer11}. Thus,
the crowdsourcer is incentivized to determine the reliability of
individual users in a scalable fashion.


This paper is organized as follows.
The iterative, unsupervised framework  and assumptions of \cite{Karger11}
in Section \ref{karger11-sec}.
In Section \ref{reg-sec}, we find expressions for the
mean and variance of the parameters ($y$) used to weight user answers
after one iteration, under
certain assumptions related to the regularity
of connectivity of the bipartite graph matching users to questions/sub-tasks.
We also state the existence of a fixed point for a normalized version
of the user-weights iteration.  To derive an
asymptotic result, we consider the user weight iteration
spanning a sequence of
independent and identically distributed (i.i.d.) meta-tasks,
with one iteration per meta-task.
We give the results of  a
numerical study for the original
system (multiple iterations for single meta-task)
in Section \ref{num-sec}.
In Section \ref{LDPC-sec}, how the crowdsourcing
system of \cite{Karger11} is related to LDPC decoding
is decribed.
Finally, we conclude with a summary in Section \ref{concl-sec}.

\section{Model Background}\label{karger11-sec}

In \cite{Karger11}, a single meta-task is divided into
a group of $|Q|$ similar subtasks/questions $i\in Q$ for which the
true Boolean answers are encoded $z_{i}\in\{-1,1\}$.
These questions are assigned to a group of $U$ users $a\in U$.
If $a$ is assigned question $i$, then his/her answer is $A_{ia}\in\{-1,1\}$.
Again, the questions $i$ are assumed similar so we model user $a$
with a task-independent parameter $p_a$ which reflects the reliability
of the user's answer: for all $i$,
\beqa
\P(A_{ia}=z_i)  =  p_a &\mbox{and} &
\P(A_{ia}=-z_i)  =  1-p_a,
\eeqa
so that
\beqa
\E A_{ia} & = & z_ip_a-z_i(1-p_a) ~=~ z_i(2p_a-1) ~~\mbox{and}\\
\var(A_{ia}) & = & 1-(2p_a-1)^2 ~=~ 4p_a(1-p_a).
\eeqa
Suppose that the  response to question $i$ is
determined by the crowdsourcer as
\be\label{hatz}
\hat{z}_i & = & \mbox{sgn}\left(\sum_{b\in\partial i} A_{ib} y_{b\ra i}\right),
\ee
where $\partial i \subset U$ is the group of users assigned
to question $i$ and $y_{b\ra i}$ is the {\em weight} given
to user $b$ for question $i$.

If $y_{b\ra i}$ is the same positive constant
for all $b,i$, then the crowdsourcer is simply taking a majority vote
without any knowledge of the reliability of the peers.

One approach to determining weights $y$
is to assess how
each user $a$ performs with respect to
the majority of those assigned to the same question $i$.
The presumption is that the majority will tend to be correct on average.
Given that, how can the crowdsourcer identify the unreliable
users/respondents so as to avoid them  for
subsequent tasks?
Accordingly, a different weight $y_{i\rightarrow a}$ can be
iteratively determined for each user $a$'s response to every
question $i$ in
the following way \cite{Karger11}:
\begin{itemize}
\item Initialize i.i.d.  $y^{(0)}_{a\rightarrow i}\sim \mbox{N}(1,1)$, \ie
initially assume each user is roughly reliable with $\E y^{(0)}=1$ and
$\P(y^{(0)}>0) \approx 0.84$.
\item For step $k\geq 1$:
\begin{itemize}
\item[$k$.1:]
$x^{(k)}_{j\ra a} =
\sum_{b\in\partial j\backslash a} A_{jb} y^{(k-1)}_{b\ra j}$,
\ie consider the weighted answer to question $j$
not including user $a$'s response.
\item[$k.2$:]
$y^{(k)}_{a\ra i} =
\sum_{j\in\partial^{-1} a\backslash i} A_{ja} x^{(k)}_{j\ra a}$,
\ie correlate the responses of the other users with
those of user $a$ over all questions assigned to $a$ except $i$.
\end{itemize}
\end{itemize}
Here $\partial^{-1} a$ is the set of questions assigned to user $a$.
The distribution of $y^{(k)}_{a\ra i}$ as a function of iteration $k$
is studied in \cite{Karger11} for degree-regular assignment of
questions to users.
Note that by simply eliminating
$x^{(k)}_{j\ra a}$ we can write
\be
y^{(k)}_{a\ra i} & = &
\sum_{j\in\partial^{-1} a\backslash i}
\sum_{b\in\partial j\backslash a} A_{ja} A_{jb} y^{(k-1)}_{b\ra j} .
\label{y-iter-karger11}
\ee
So, $y^{(1)}_{a\ra i}$ depends on the responses of $a$'s
{\em one-hop neighbors} in $U$,
\beqa
N^{(1)}_{a\ra i} & := &
\{b \in U ~|~ \exists j\in \partial^{-1} a \backslash i ~
\mbox{s.t.}~b\in \partial j\backslash a\},
\eeqa
\ie not including $a$ itself or any one-hop
neighbors of $a$ (in $U$) also assigned to question $i$.

\section{Distribution of user weights after one iteration}\label{reg-sec}

\subsection{Degree-regular graph}\label{deg-reg-sec}
For the degree-regular assignment,
we can relate the number of users per question $r:=|\partial i|~\forall i$ to
the number of questions per user $s:=|\partial^{-1} a|~\forall a$:
\beqa
r|Q| & = & s|U|.
\eeqa
In the following, we will assume
\beqa
r\geq 2 & \mbox{and} & s \geq 2.
\eeqa
Furthermore, we may assume that
all sets $N^{(1)}_{a\ra i}$
are the same size $N$, where
generally
\beqa
N & \leq & (r-1)(s-1)~\mbox{members},
\eeqa
but
that the number of terms summed in
(\ref{y-iter-karger11}) is
always equal to $(r-1)(s-1)$.

To form such degree-regular assignments, one
can simply iterate over the (enumerated) questions:
\begin{enumerate}
\item[0.] $i=1$ (first question $\in\{1,2,...,|Q|\}$).
\item[1.] assign $i$ to $r$ different users $\in U$ chosen
uniformly at random.
\item[2.] $\forall a\in U$ such that $|\partial^{-1}a|=s$,
$U\rightarrow U\backslash \{a\}$.
\item[3.] if $i<|Q|$, $i\rightarrow i+1$ and go to step 1.
\end{enumerate}
Note that since $r|Q|=s|U|$, the questions will
be exhausted just when the users are (\ie when
$i\rightarrow |Q|$, $U\rightarrow \emptyset$).

\subsection{First-iteration variance and mean of user weights, $y$}

Let
$O^{(0)}_{ja}  =  \sum_{b\in\partial j\backslash a} A_{jb} y^{(0)}_{b\ra j}$
and note that $O^{(0)}_{ja}$ is independent of
$O^{(0)}_{j'a}$ for all $a$ and $j\not = j'$.
So, if $ \mu^{(0)}_{a\ra i} := \E y^{(0)}_{a\ra i} =1 $,
the mean of $y^{(1)}_{a\ra i} $ is
\beqa
\mu^{(1)}_{a\ra i} & := &
\sum_{j\in\partial^{-1} a\backslash i} \E A_{ja}
\sum_{b\in\partial j\backslash a} \E A_{jb} \cdot 1
~~\mbox{(by indep.)}  \\
& = & \sum_{j\in\partial^{-1} a\backslash i} z_j (2p_a-1)
\sum_{b\in\partial j\backslash a} z_j (2p_b-1)  \\
& = & (2p_a-1)
\sum_{j\in\partial^{-1} a\backslash i}
\sum_{b\in\partial j\backslash a}  (2p_b-1)
\eeqa
(since $z_j^2 =1$ a.s.).
Also, assume the variance $\var^{(0)}_{a\ra i}:=
\var(y^{(0)}_{a\ra i}) =1 $
($\Rightarrow \E(y^{(0)}_{a\ra i})^2 =2 $).
So,
\be
\var^{(1)}_{a\ra i}& = &
\sum_{j\in\partial^{-1} a\backslash i} \var( A_{ja} O^{(0)}_{ja} )
~~\mbox{(by indep.)}
\nonumber\\
& = &
\sum_{j\in\partial^{-1} a\backslash i} [
\E(O^{(0)}_{ja})^2 -
(2p_a-1)^2(\E O^{(0)}_{ja})^2]
\nonumber\\
& = &
\sum_{j\in\partial^{-1} a\backslash i} [
\var(O^{(0)}_{ja})^2 +
(1-(2p_a-1)^2)(\E O^{(0)}_{ja})^2]
\nonumber\\
& = &
\sum_{j\in\partial^{-1} a\backslash i} [
\{\sum_{b\in\partial j\backslash a}
 \var(A_{jb} y^{(0)}_{b\ra j})\}+ \nonumber \\
& & ~(1-(2p_a-1)^2)(\E O^{(0)}_{ja})^2]
~~\mbox{(by indep.)}
\label{var1-indep}\\
& = &
\sum_{j\in\partial^{-1} a\backslash i} [
\{\sum_{b\in\partial j\backslash a}
 (2-(2p_b-1)^2)\}+\nonumber\\
& &
~(1-(2p_a-1)^2)(\sum_{b\in\partial j\backslash a}
2p_b-1)^2].
\label{y-iter-karger-var1}
\ee

\subsection{Assumption of large number of users per question, $r$}

Finally, for simplicity,
we may additionally assume sufficiently large $r$
(number of users per question) and the neighbor
selection is uniformly distributed so that, for all $a,j$,
\be\label{r-assumption}
\frac{1}{r-1}\sum_{b\in\partial j\backslash a} (2p_b-1)
 & \approx & \E (2p -1) ~ = ~ 2\E p -1.
\ee

The following lemma is now obtained simply by substitution.

\begin{lemma}
For (\ref{y-iter-karger11}) under (\ref{r-assumption}), for all $a,i$:
\be
\mu^{(1)}_{a\ra i}
 & \approx & (s-1)(r-1) (2p_a-1)(2\E p -1)  \label{mu1}
\ee
and
\be
\var^{(1)}_{a\ra i}& \approx & (s-1)(r-1)[\E(2-(2p-1)^2) \nonumber\\
& & ~ + (1-(2p_a-1)^2) (r-1) (2\E p-1)^2 ].
\label{var1}
\ee
\end{lemma}

Though the $U\times Q$ matrix
$\bfY^{(k)}$
with elements
$Y_{a,i}:=y^{(k)}_{a\ra i}$, is Markovian,
directly proceeding along these lines for
$\var^{(k)}_{a\ra i}$, $k\geq 2$, is complicated by
the dependence of the terms involved
through the structure of the bipartite graph
mapping users $U$ to questions $Q$,
\cf Section \ref{unsup-series-sec}.


\subsection{Discussion: Normalized weights}\label{unsup-norm-sec}

It's possible that the weights $y^{(k)}$
may be unbounded in $k$.
Instead of (\ref{y-iter-karger11}), for a
degree-regular assignment suppose the weights are,
for all $a,i$,
\be
\hy^{(k)}_{a\ra i} & = &  \frac{1}{(s-1)(r-1)}
\sum_{j\in\partial^{-1} a\backslash i}
\sum_{b\in\partial j\backslash a} A_{ja} A_{jb} \hy^{(k-1)}_{b\ra j}.
\label{y-iter-norm}
\ee
Let $\bfhY^{(k)}$ be the $|U|\times |Q|$-matrix with elements
$\hY_{a,i} := \hy^{(k)}_{a\ra i}$.

\begin{proposition}
For (\ref{y-iter-norm}),
if $\bfhY^{(0)} \in [-1,1]^{|U|\times|Q|}$, then
the sequence $\bfhY^{(k)}$
has a fixed point in
$[-1,1]^{|U|\times|Q|}$.
\end{proposition}
\IEEEproof
Simply by the triangle inequality and induction, if
$\bfhY^{(0)} \in [-1,1]^{|U|\times|Q|}$ then
$\bfhY^{(k)} \in [-1,1]^{|U|\times|Q|}$ for all $k$.
As the mapping
(\ref{y-iter-norm}) is continuous, we can
apply Brouwer's fixed point theorem \cite{Border85}
to get existence.
\qed

~\\

Note that, generally,  fixed points of a continuous linear operator
on a bounded domain needn't be unique.

\section{A series of similar meta-tasks with one iteration per
meta-task}\label{unsup-series-sec}


Let
\beqa
\delta  & := & (s-1)(r-1) ~> 1, ~\mbox{and}\\
\phi & := & \E(2p-1)^2 ~\in[0,1].
\eeqa
We now consider a {\em series} of similar meta-tasks
indexed $k$ and a {\em single} iteration
as (\ref{y-iter-karger11}) for each on its component questions
(all questions similar to each other too).
Moreover, each meta-task
will reassign the component questions using an independently
sampled degree-regular assignment. Obviously, the answers $A$ will
be independently resampled too. That is, here
\be\label{y-iter-series}
\ty^{(k)}_{a\ra i} & = &
\frac{1}{\delta}
\sum_{j\in\partial^{-1} a^{(k)}\backslash i}
\sum_{b\in\partial j^{(k)}\backslash a}
A_{ja}^{(k)}
A_{jb}^{(k)} \ty^{(k-1)}_{b\ra j}
\ee
{\em where now the
$A_{jb}^{(k)}$ and  $\ty^{(k-1)}_{b\ra j}$ terms are independent}.
The following asymptotic analysis is facilitated
by this assumption on successive i.i.d. meta-tasks.

\begin{lemma}\label{mu-lemma}
If  (\ref{r-assumption}) and $\tmu^{(0)}_{a\ra i}=1$
for all $a,i$, then for all $k\geq 1$,
\be\label{muk}
\tmu^{(k)}_{a\ra i}
 & \approx & (2p_a-1) (2\E p -1)\phi^{k-1}.
\ee
\end{lemma}

\IEEEproof
First note that by the argument
for (\ref{mu1})  and definition (\ref{y-iter-series}),
(\ref{muk}) holds for $k=1$.
The lemma is simply proven by induction.
\qed

~\\
\begin{lemma}\label{var-lemma}
If  (\ref{r-assumption}),
$\tmu^{(0)}_{a\ra i}=1$
and $\tvar^{(0)}_{a\ra i}=v_0$
for all $a,i$, then for $k\geq 1$
\be \label{vark}
\var^{(k)}_{a\ra i} & \leq &
v_0\delta^{-k}
+r(2\E p -1)^2\delta^{-1}\frac{\phi^{2k}-\delta^{-k}}{\phi^2-\delta^{-1}}.
\ee

\end{lemma}

\IEEEproof
Proceeding as for (\ref{y-iter-karger-var1}),
and using the independence
at (\ref{var1-indep}) afforded by (\ref{y-iter-series}),
gives
\beqa
\tvar^{(k)}_{a\ra i}
& = &
\frac{1}{\delta^2}
\sum_{j\in\partial^{-1} a^{(k)}\backslash i} [
\{\sum_{b\in\partial j^{(k)}\backslash a}
 \var(A_{jb}^{(k)} \ty^{(k-1)}_{b\ra j})\}\\
& & ~+ (1-(2p_a-1)^2)\{
\sum_{b\in\partial j^{(k)}\backslash a}
(2p_b-1)\tmu^{(k-1)}_{b\ra j}
\}^2]\\
& = &
\frac{1}{\delta^2}
\sum_{j\in\partial^{-1} a^{(k)}\backslash i} [
\{\sum_{b\in\partial j^{(k)}\backslash a}
 \E(\ty^{(k-1)}_{b\ra j})^2\\
& & ~~~ - (2p_b-1)^2 (\tmu^{(k-1)}_{b\ra j})^2\}\\
&&~+
(1-(2p_a-1)^2)\{
\sum_{b\in\partial j^{(k)}\backslash a}
(2p_b-1)\tmu^{(k-1)}_{b\ra j}
\}^2]\\
& = &
\frac{1}{\delta^2}
\sum_{j\in\partial^{-1} a^{(k)}\backslash i} [
\{\sum_{b\in\partial j^{(k)}\backslash a}
 \tvar^{(k-1)}_{b\ra j}\\
& & ~ + (1- (2p_b-1)^2) (\tmu^{(k-1)}_{b\ra j})^2\}\\
& & ~+
(1-(2p_a-1)^2)\{
\sum_{b\in\partial j^{(k)}\backslash a}
(2p_b-1)\tmu^{(k-1)}_{b\ra j}
\}^2]\\
 & \leq &
\sum_{j\in\partial^{-1} a^{(k)}\backslash i} [
\{\sum_{b\in\partial j^{(k)}\backslash a}
 \tvar^{(k-1)}_{b\ra j} + (\tmu^{(k-1)}_{b\ra j})^2\}\\
& & ~+
\{
\sum_{b\in\partial j^{(k)}\backslash a}
(2p_b-1)\tmu^{(k-1)}_{b\ra j}
\}^2].
\eeqa
Thus,
\beqa
\tvar^{(k)}_{a\ra i}
& \leq &
\frac{1}{\delta^2}
\sum_{j\in\partial^{-1} a^{(k)}\backslash i}
\sum_{b\in\partial j^{(k)}\backslash a}
 \tvar^{(k-1)}_{b\ra j} \\
& & ~
+ \frac{(2\E p -1)^2(\E(2p-1)^2)^{2k}}{\delta}\\
& & ~
+\frac{(r-1)(2\E p -1)^2(\E(2p-1)^2)^{2k}}{\delta}\\
& = &
\frac{1}{\delta^2}
\sum_{j\in\partial^{-1} a^{(k)}\backslash i}
\sum_{b\in\partial j^{(k)}\backslash a}
 \tvar^{(k-1)}_{b\ra j}\\
& & ~
+ \frac{r(2\E p -1)^2 \phi^{2k}}{\delta}.
\eeqa
The proof then follows by induction, \ie dropping
dependence on $a,i$, the previous display is
\beqa
\tvar^{(k)} & \leq &
 \frac{1}{\delta}\tvar^{(k-1)}
+ \frac{1}{\delta}r(2\E p -1)^2\phi^{2k}.
\eeqa

\qed

~\\

By direct substitution, we arrive at the following.

\begin{proposition}
For (\ref{y-iter-series})\footnote{With or without the
normalizing factor $\tfrac{1}{\delta}$.}
under (\ref{r-assumption}),
if
\be
\frac{1}{(s-1)(r-1)} & \leq & [\E(2p-1)^2]^2 ~<~1, \label{delta-phi-cond}
\ee
then for all $a,i$ with $p_a\not = 0.5$,
\beqa
\frac{\sqrt{\tvar^{(k)}_{a\ra i}}}{\tmu^{(k)}_{a\ra i}} & = &
\mbox{O}(1),
\eeqa
\ie the relative error
is bounded as $k\rightarrow \infty$.
\end{proposition}

Note that (\ref{delta-phi-cond}) is just $\delta^{-1} \leq  \phi^2    < 1$.



In (\ref{muk}),
the product $(2p_a-1)(2\E p -1)$ determines the {\em sign} of
$\mu^{(k)}_{a\ra i}$ (corresponding to the weight of user $a$ for question
$i$). So,  if the crowd
tends to be correct ($\E p > 0.5$) then (as expected):
if  a particular user $a$ tends to be correct ($p_a > 0.5$)
then the sign of $a$'s weight $y_{a\rightarrow i}$ will tend
to be positive, else negative.

Revisiting
(\ref{hatz}), we can instead estimate the answer to question $i$
of task $k$ using {\em normalized} weights as
\beqa
\hat{z}^{(k)}_i & = & \mbox{sgn}\left(\sum_{b\in\partial i} A_{ib}^{(k)}
\ty^{(k)}_{b\ra i}
\right)
\eeqa
An immediate consequence of the previous proposition
 is the following (noting $\phi$ in (\ref{muk}) is positive).

\begin{corollary}\label{small-rel-err-cor}
If the limiting relative error is sufficiently small,
then
\be\label{decision}
\hat{z}^{(k)}_i & \sim &
\mbox{{\em sgn}}\left((2\E p -1)\sum_{b\in\partial i} A_{ib}^{(k)}
(2p_b-1) \right)
\ee
as $k\rightarrow\infty$.
\end{corollary}

~\\
This expression is intuitive pleasing as the
reliability of the individual user $b$'s responses
$A_{ib}$
are weighted by their own reliabilities $(2p_b-1)$ essentially
learned through the iterations.
Also, the overall reliability of the crowd, $\E (2p-1)=2\E p -1$, is a factor
so that if the crowd is on average unreliable
($\E(2 p-1)<0$) the opposite response (sign change) of the system which favors
the majority (the summation) will result.

%
%
%

\section{Numerical Study}\label{num-sec}

Considering the term $\phi:=\E(2p-1)^2$ in (\ref{delta-phi-cond}),
note that  the crowd is unreliable
if $\E p$ is close to 0.5. If
$\E(2p-1)^2\not = 0$ but is small, then  intuitively
the unreliability of the crowd can
can be compensated for by sufficient correlation
with the majority, \eg if (\ref{delta-phi-cond}) holds
and a sufficient number of iterations $k$ are performed,
otherwise the weights  $\bfY$ (or $\bftY$) may be too close to
${\bf 0}$ giving indeterminant decisions (\ref{hatz}).

\begin{figure}[!t]
\begin{center}
\includegraphics[width=3in]{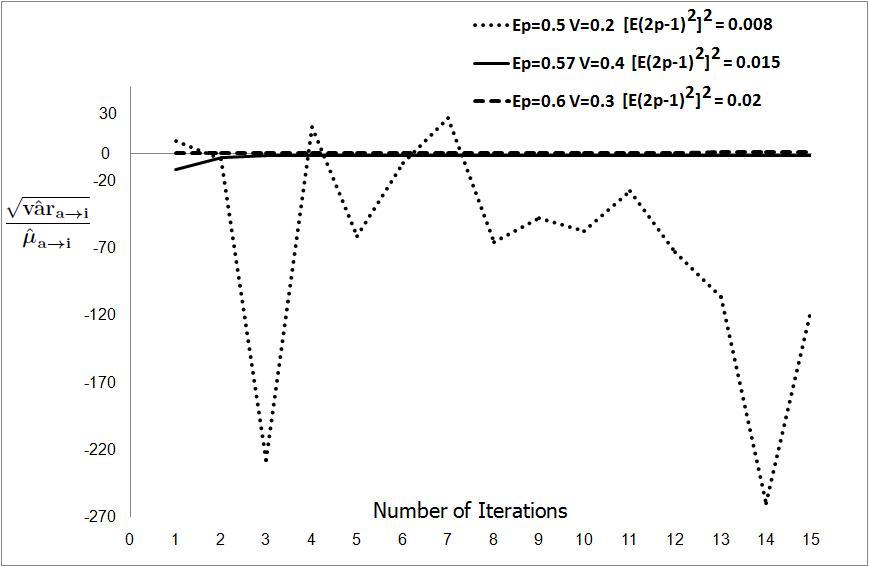}
\caption{Relative error $\bf{\frac{\sqrt{\hat{var}_{a\ra i}}}{\hat{\mu}_{a\ra i}}}$ with iterations}\label{sim1}
\vspace{-0.25 in}
\end{center}
\end{figure}

A model of a crowdsourcing assignment with {\em single} meta-task was simulated using Netlogo \cite{Netlogo}, a multi-agent simulation tool.
The meta-task was an aggregate of $100$ sub-tasks/questions to be assigned to a subset of $100$ users.
Using the method described in Section \ref{deg-reg-sec}, we performed degree regular question-to-user assignment. Each user was assigned $10$ questions, \ie $s=r=10$.
We generated random reliabilities $p_a$ for each user $a$ using normal distribution with a known mean $\E p$ and variance
{\sf V}$=\E(p^2) - (\E p)^2$.
Using these reliabilities, random answers were generated by each user for the questions assigned.
The weights $y_{a \ra i}$ for each link between user $a$ and question $i$ were randomly initialized with normal distribution with mean and variance equal to $1$ as in \cite{Karger11}.
We computed the values of $y_{a \ra i}$ for all $a$ and $i$
according to (\ref{y-iter-karger11})
for $k=15$ iterations of message passing between the users and the questions and vice-versa.
For each value of $\E p$ and {\sf V} we repeated the previous step (consisting of $15$ iterations of message passing) $50$ times, each time generating new answers $A_{ia}$ for all $i$ and $a$ based on the reliabilities of the users.
Figure \ref{sim1}  is a plot for a (typical) link $(a,i)$ of
$\bf{\frac{\sqrt{\hat{var}_{a\boldsymbol{\ra} i}}}{\hat{\boldsymbol{\mu}}_{a\boldsymbol{\ra} i}}}$
versus iteration index ($k$), for different values of $\E p$ and {\sf V},
where $\hat{\mu}_{a\ra i}$ and $\hat{\var}_{a\ra i}$ are the sample mean and sample variances respectively of the weights of edge $(a,i)$ for a given iteration.
Note that  $\frac{1}{(s-1)(r-1)}= \frac{1}{81}=0.0123$ and when $\E p =0.5$ (\ie an unreliable group of users), $[\E(2p-1)^2]^2 = 0.008 < 0.0123$,
there is no convergence, otherwise there is rapid convergence
of the relative error to zero, so that the condition of
Corollary \ref{small-rel-err-cor} is met for this single meta-task
experiment.

\begin{figure}[!t]
\begin{center}
\includegraphics[width=3in]{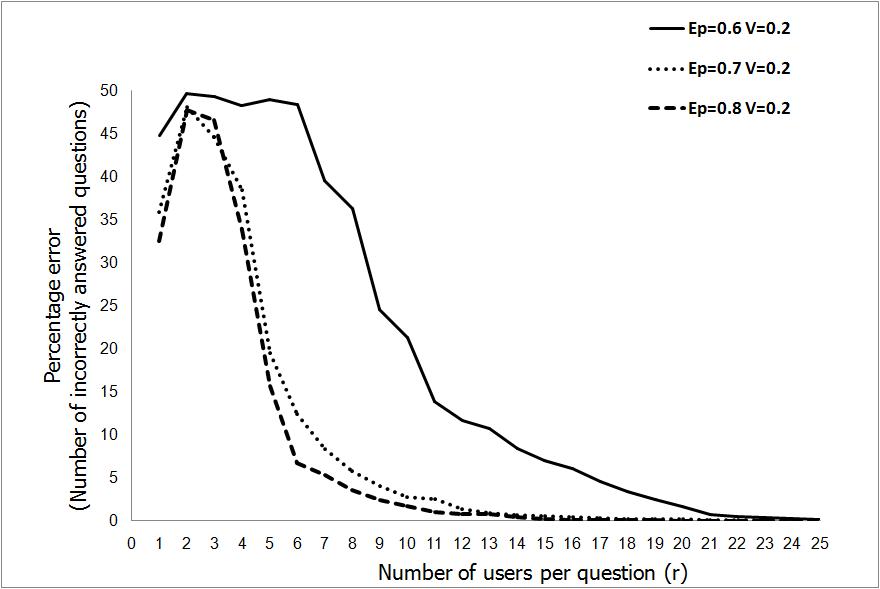}
\caption{Effect of $r$ on the percentage error for different user reliabilities}\label{sim2}
\vspace{-0.25 in}
\end{center}
\end{figure}

In our next experiment we varied $r$ \ie the number of users assigned to a question and observed the average percentage error (the number of questions with incorrect answers derived through the weighted majority correlation method). The error values were steady state values \ie when the iterative calculation of weights $y_{a \ra i}$ converged and we took the weighted correlation with the majority of users. The average was taken for $50$ different random realizations of the question-to-user assignment and answers for the given value of $r$ , $\E_p$ and {\sf V}. Fig \ref{sim2} shows the percentage error for different values of $\E p$ as it decreases with $r$. Note that for all of these three pairs of $\E_p$ and {\sf V}, the value of $[\E(2p-1)^2]^2$ was well above $\frac{1}{(s-1)(r-1)}$. We observed an initial increase in the error approximately at $r=2,3$. A possible explanation for this could be the labeling of reliable users as unreliable (by lowering the weight $y_{a \ra i}$ for the user) in the light of insufficient samples to obtain correct correlation.

\section{Similarity with LDPC Decoding via Message Passing}\label{LDPC-sec}

The consensus framework described above has a marked resemblance with message passing algorithms of belief propagation for decoding of low-density-parity-check (LDPC) codes \cite{LDPCtutorial04}.
LDPC codes were introduced in early sixties \cite{Gallager}, but became popular only recently because of their ability to reach the Shannon limits on communication rates. LDPC codes typically have large lengths and rely on message passing over sparse bipartite graphs to reach a consensus on the transmitted codeword.
One can think of leveraging the already mature theory of LDPC coding and decoding for the crowdsourcing model.
For instance, the convergence properties of LDPC codes are normally studied under an independence assumption, \ie that
the messages received by each node are independent.
This assumption is true for the first few iterations defined by the ``girth length" of the codes, which is the length of the smallest cycle in the graph.
In our model, girth length will depend on the question-to-user assignment.

Consider a model where the weights $y_{a \ra i} \in [0,1]$
because instead of (\ref{y-iter-karger11}),
\be \nonumber
y^{(k)}_{a\ra i} =
\frac{1}{s-1}\sum_{j\in\partial^{-1} a\backslash i}\mbox{sgn}^{+} \left(A_{ja}~ \mbox{sgn} \left(
\sum_{b\in\partial j\backslash a}  A_{jb} y^{(k-1)}_{b\ra j} \right) \right),
\label{y-iter-karger11-ldpc}
\ee
where $\mbox{sgn}^{+}(a)=1$ if $a>0$ and $0$ otherwise. Here, $x_{j \rightarrow b}$ gives the answer based on the expected value of answers given by all users $a\in\partial j\backslash b$, and  $y_{a \rightarrow i}$ is the posterior probability that $a$ answers $i$ correctly given that we know the correct answers for all questions $j\in\partial^{-1} a\backslash i$.
Alternatively, soft decisions can be used, \eg by defining $x_{j \rightarrow b}$ as the log-likelihood of the answer being $0$ or $1$.
The question nodes compute the user-reliability expectations, while the user nodes maximize the log-likelihood of the observed answers over their (estimated) reliabilities.
So, this is similar to the Expectation-Maximization (EM) algorithm \cite{EM-algo}. 

The use of EM algorithm is natural in this scenario since we have to estimate a set of decision variables (correct answers) along with latent variables (the user reliabilities) \cite{EM-1}. Apart from the user reliabilites, one can also think of considering difficulty of the tasks as another set of latent variables \cite{EM-2}. The M-step of the EM algorithm poses some computational challenge and most of the known work in this area seeks to find a solution to the maximization step by using softwares that use numerical optimization techniques. From this perspective the message passing framework can be viewed as an alternate local or distributed optimization technique that takes node-by-node decisions iteratively. It will be interesting to find the cost of using such a framework in terms of the loss in optimality.

\section{Summary}\label{concl-sec}

This paper studied iterative, unsupervised crowdsourcing
frameworks wherein the weights of users' answers are
determined by correlating their responses to the majority.
We considered the case of a
single meta-task and multiple independent meta-tasks,
deriving an asymptotic result for the latter.
Numerical experiments for multiple iterations on
a single meta-task show that the iteration does not
converge when the the crowd is unreliable, but rapid
convergence otherwise results.
Finally, we briefly described how these crowdsourcing
frameworks are related to LDPC decoding and EM.


\begin{thebibliography}{99}
\bibitem{Mec}
Amazon Mechanical Turk. http://www.mturk.com

\bibitem{Bai99}
Z.D. Bai.
Methodologies in spectral analysis of large
dimensional random matrices, a review.
{\em Statistica Sinica} {\bf 9}:611-677,
1999.

\bibitem{Border85}
K.C. Border.  Fixed Point Theorems with Applications to Economics and
        Game Theory.
Cambridge University Press, London, 1985.


\bibitem{BerkeleySeminar}
Crowdsourcing Seminar.
http://husk.eecs.berkeley.edu/courses/cs298-52-sp11/index.php/Main\_Page

\bibitem{BLM10}
M.S. Bernstein, G. Little, R.C. Miller, B. Hartmann, M.S. Ackerman,
D.R. Karger, D. Crowell, and K. Panovich.
Soylent: A word processor with a crowd inside.
In {\em Proc. ACM Symp. User Interface Software and Tech.},  New York, NY,
2010.

\bibitem{Cas}
Casting Words. http://castingwords.com

\bibitem{Cro}
Crowd Flower. http://crowdflower.com

\bibitem{EM-1}
A.P. Dawid and A.M. Skene. Maximum likelihood estimation of observer
error-rates using the EM algorithm.
{\em J. Royal Stat. Soc. C} {\bf 28}(1):20-28, 1979.

\bibitem{Gallager}
G. Gallager.
Low Density Parity-Check Codes.
In {\em MIT Press}, Cambridge, MA. 1963.

\bibitem{GJ09}
B. Golub and M.O. Jackson.
How homophily affects learning and diffusion in networks.
Feb. 2009, available at
http://arxiv.org/PS\_cache/arxiv/pdf/0811/0811.4013v2.pdf

\bibitem{Hoffman11}
Leah Hoffman.
Crowd Control. {\em Commun. of the ACM}
{\bf 52}(3):16-17, 2009.


\bibitem{IPW10}
P.G. Ipeirotis, F. Provost, and J. Wang.
Quality management on Amazon Mechanical Turk.
In {\em Proc. ACM SIGKDD Workshop on Human Computation},
New York, NY, 2010.

\bibitem{Karger11}
D.R. Karger, S. Oh and D. Shah.
Iterative Learning from a Crowd (abstract).
In {\em Proc. Interdisciplinary Workshop on
Information and Decision in Social Networks},
MIT, May 31 - June 1, 2011.
http://wids.lids.mit.edu/wids\_program\_final.pdf p.32-33.



\bibitem{Kulk11}
A. Kulkarni, M. Can, B. Hartmann.
{\em Turkomatic: Automatic, Recursive Task and Workflow Design for
Mechanical Turk}. (Poster)
In {\em Proc. AAAI Human Computation Workshop (HCOMP)}, 2011.


\bibitem{MP}
Marchenko-Pastur Distribution. Available at\\
http://en.wikipedia.org/wiki/Marchenko-Pastur\_law

\bibitem{Netlogo}
NetLogo itself: Wilensky, U. 1999. NetLogo.
\newblock http://ccl.northwestern.edu/netlogo/. Center for Connected Learning and Computer-Based Modeling, Northwestern University. Evanston, IL.

\bibitem{Savage10}
M. Motoyama, K. Levchenko, C. Kanich, D. McCoy, G.M. Voelker and S. Savage.
Understanding CAPTCHA-Solving from an Economic Context.
In {\em Proc. USENIX Security Symposium}, Washington, DC, August 2010.

\bibitem{EM-3}
V. Raykar, S. Yu, L. Zhao, G. Valadez, C. Florin, L. Bogoni, and L. Moy.
Learning from Crowds.
In {\em Journal of Machine Learning Research}, 11(7):1297–1322, 2010.

\bibitem{Singer11}
Y. Singer and M. Mittal.
Pricing Mechanisms for Online Labor Markets.
In {\em Proc. AAAI Human Computation Workshop (HCOMP)}, 2011.

\bibitem{Soy}
Soylent.  http://projects.csail.mit.edu/soylent/

\bibitem{LDPCtutorial04}
A. Shokrollahi.
LDPC codes: An introduction.
In {\em Coding, cryptography and combinatorics, Progress in Computer Science and Applied Logic}, 2004

\bibitem{EM-algo}
 C.  Tomasi.
 Estimating Gaussian Mixture Densities with EM – A Tutorial.
 http://citeseer.nj.nec.com


\bibitem{EM-2}
J. Whitehill, P. Ruvolo, T. Wu, J. Bergsma, and J. Movellan. 
Whose vote should count more: Optimal integration of labels from labelers of unknown expertise.
In {\em NIPS}, 22:2035–2043, 2009.

\end{thebibliography}
\end{document}